\documentclass[11pt]{article}

\usepackage[utf8]{inputenc}
\usepackage[english]{babel}
\usepackage[margin=1in]{geometry}
\usepackage{times}
\usepackage{microtype}
\usepackage{amsmath}
\usepackage{amssymb}
\usepackage{graphicx}
\usepackage{booktabs}
\usepackage{multirow}
\usepackage{array}
\usepackage{enumitem}
\usepackage{url}
\usepackage{hyperref}
\usepackage[numbers]{natbib}
\usepackage{tikz}
\usepackage{pgfplots}
\usepgfplotslibrary{groupplots}
\usetikzlibrary{positioning,arrows.meta,calc,fit,shapes.geometric}
\pgfplotsset{compat=1.18}

\hypersetup{
  colorlinks=true,
  linkcolor=black,
  citecolor=black,
  urlcolor=blue,
  pdftitle={StateFuse: Deterministic Conflict-Preserving Memory for Multi-Agent Systems},
  pdfauthor={Sergey Volkov, Yang Li, and Ye Luo}
}

\newcommand{\RepoURL}{https://github.com/nZiben/statefuse}
\newcommand{\keywords}[1]{%
  \par\noindent\textbf{Keywords: }#1\par\vspace{0.8em}%
}
\providecommand{\doi}[1]{doi: #1}
\title{StateFuse: Deterministic Conflict-Preserving Memory for Multi-Agent Systems}
\author{
  Sergey Volkov\\
  The University of Hong Kong\\
  \texttt{savolkov@hku.hk}
  \and
  Yang Li\\
  The University of Hong Kong\\
  \texttt{yli9919@hku.hk}
  \and
  Ye Luo\\
  The University of Hong Kong\\
  \texttt{kurtluo@hku.hk}
}
\date{}

\begin{document}

\maketitle

\begin{abstract}
Agent systems accumulate conflicting observations across branches, retries, and replicas, yet many practical memory layers still collapse disagreement behind overwrite rules that are difficult to inspect or correct.
We present \textit{StateFuse}, a conflict-aware replicated memory contract built on standard OpSet/CRDT merge.
StateFuse does not introduce a new join algebra; it defines an agent-facing semantics layer with immutable history, explicit conflict objects, exact and semantic correction handles (\texttt{claim\_id} / \texttt{claim\_ref}), deterministic predicate contracts, and projection-time resolution that cannot rewrite replicated state.

We evaluate StateFuse against flat multi-value, raw-log, provenance-style, and collapsed baselines under matched resolver and verification policies.
On a 282-question official conflict-bearing MemoryAgentBench slice, the compared methods tie on answer accuracy, but conflict-preserving surfaces keep contradictions visible while collapsed surfaces do not.
In a controlled agent loop with uniform verification, preserving ambiguity enables safer abstention and correction than early collapse.
A correction-handle ablation further shows that semantic handles matter when exact prior identifiers are unavailable.

The resulting claim is narrow:
StateFuse is best supported as a safer public memory contract for contradiction surfacing, abstention, and auditable correction, not as a universal accuracy gain.
\end{abstract}


\begin{center}
{\small Code and supplementary materials: \url{\RepoURL}}
\end{center}

\section{Introduction}
\label{sec:intro}

Agent systems increasingly operate as branching, tool-using, distributed processes.
They speculate, retry, fork into subplans, and later reconcile what each branch believes.
In this setting, memory is not only a retrieval problem but also a \emph{correction and merge semantics} problem:
when two branches assert incompatible facts about the same entity, should the system overwrite one value, preserve both, retract one explicitly, or defer the choice to task-local policy?
Many practical memory stacks leave this question implicit.
The result is a familiar cluster of failures:
hidden overwrite of branch disagreement, stale observations surviving validated corrections, and unsafe actions taken from collapsed memory surfaces.

Distributed systems already provide the right convergence substrate.
CRDT joins and OpSet-style interpretations show how replicas can exchange immutable updates and recover equivalent state through associative, commutative, and idempotent merge \citep{almeida2018delta,Kleppmann2018OpSets,kleppmann2019localfirst,shapiro2011comprehensive}.
StateFuse does \emph{not} propose a new join.
Its focus is the contract that sits on top of that substrate for agent memory:
what the replicated history stores, what a public projection must expose, how corrections are targeted, and what downstream policies are allowed to decide.

We make two design choices in the evaluation.
First, we narrow the empirical claim.
We do not argue that explicit conflict objects universally beat strong flat multi-value surfaces on answer accuracy.
Instead, we argue that StateFuse provides a stronger public memory contract for contradiction surfacing, abstention, and auditable correction.
Second, we make the main comparisons fairer.
Flat multi-value and raw-log baselines now receive conservative variants from the same resolver family, and the downstream agent loop gives every method the same verification budget.

The StateFuse contract has five main components:
\begin{enumerate}[leftmargin=*]
  \item \textbf{Immutable replicated history} for evidence, claims, retractions, and decisions.
  \item \textbf{Explicit public conflict surfacing} at projection time rather than silent overwrite at merge time.
  \item \textbf{Dual correction handles}: exact \texttt{claim\_id} for local edits and semantic \texttt{claim\_ref} for cross-replica or previously unseen targets.
  \item \textbf{Deterministic predicate contracts} for normalization, equality, and claim-reference derivation.
  \item \textbf{Bounded projection authority}: resolvers may choose among surfaced candidates or abstain, but they cannot rewrite replicated state.
\end{enumerate}

Secondary engineering mechanisms such as compaction and authenticated sync remain outside the headline empirical claim.
The main paper evidence focuses on the public decision surface.

The most credible empirical question is therefore:
\begin{quote}
When matched-information baselines receive the same downstream policy budget, what does StateFuse add beyond a strong flat multi-value surface?
\end{quote}
Our answer is intentionally modest.
On the converted official MemoryAgentBench conflict-bearing slice, StateFuse and the flat conflict-preserving baselines tie on top-line accuracy, while raw-log and collapsed surfaces also tie on accuracy because the gold frequently follows the latest fact.
What differs is what the memory surface exposes:
StateFuse and flat multi-value surfaces surface contradictions on all of those tasks, whereas raw-log and collapsed surfaces surface none.
On the controlled agent loop with uniform verification, every conservative non-collapsing surface reaches the same safe operating point, while the collapsed surface remains materially worse.
That pattern says more about the value of preserving and surfacing disagreement than about any unique accuracy advantage for StateFuse itself.

Our contributions are therefore:
\begin{enumerate}[leftmargin=*]
  \item A \textbf{conflict-aware memory contract} for agent systems with immutable history, explicit conflict surfacing, exact and semantic correction handles, deterministic predicate contracts, and projection-bounded resolution.
  \item A \textbf{fairer matched-information evaluation design} in which strong flat baselines receive the same resolver family and downstream verification budget rather than weaker downstream policies.
  \item An \textbf{evaluation} centered on an official MemoryAgentBench conflict-bearing slice, a uniform-verification agent loop, and semantic-handle ablations.
\end{enumerate}

The remainder of the paper is organized as follows.
Section~\ref{sec:related} positions StateFuse relative to CRDTs, provenance-aware systems, event sourcing, and agent-memory evaluation work.
Section~\ref{sec:method} defines the contract boundary.
Section~\ref{sec:experiments} reports the main empirical evidence and semantic-handle support.
Section~\ref{sec:discussion} discusses what the evidence does and does not establish.

\section{Related Work}
\label{sec:related}

\subsection{CRDT Joins, OpSets, and Structured Local-First Replication}
StateFuse inherits its convergence substrate directly from CRDT and OpSet work \citep{almeida2018delta,Kleppmann2018OpSets,shapiro2011comprehensive}.
Merge is still set union over immutable operations, and the central claim of this paper does not depend on a new join algebra.
The closest semantic neighbors remain multi-value registers, remove-wins styles of set semantics, and structured local-first data models such as JSON CRDTs \citep{KleppmannBeresford2017JSONCRDT,kleppmann2019localfirst}.
What StateFuse adds on top is an agent-facing contract:
explicit conflict objects, semantic correction handles, projection-equivalent compaction, and an authenticated sync boundary.

Recent work on the Automerge model checker underscores that the interesting questions in replicated systems are often at the interpretation and invariant level rather than at the bare merge operator \citep{JefferyMortier2023AMC}.
This observation directly motivates our emphasis on semantic-correction, compaction, and sync invariants.

\subsection{Truth Maintenance, Belief Revision, and Provenance-Aware Data}
StateFuse is also closely related to older traditions that treat contradiction and correction as first-class objects rather than as implementation accidents.
Truth-maintenance and belief-revision systems study how conflicting beliefs should be represented, revised, and justified.
Provenance-aware data management similarly emphasizes that where a fact came from can matter as much as the fact itself.
StateFuse does not attempt to solve general belief revision.
Instead, it imports the practical lessons of those systems into a replicated agent-memory setting:
retractions are explicit, provenance remains attached to claims, and contradiction surfacing is part of the public decision interface rather than a hidden storage detail.

\subsection{Event-Sourced and Policy-at-Read-Time Systems}
Classical optimistic replication and Bayou-style weakly connected storage made application-specific conflict handling a first-class design problem \citep{SaitoShapiro2005OptimisticReplication,Terry1995Bayou}.
Event-sourcing and CQRS-style system design make a similar engineering move:
preserve append-only history and let projection code determine task-specific views.
StateFuse is philosophically close to both.
Recent workflow infrastructure such as AgentGit adds Git-like rollback and branching to multi-agent execution graphs \citep{Li2025AgentGit}.
That line of work is complementary to ours:
it focuses on checkpointed workflow control and recovery, whereas StateFuse focuses on mergeable semantic memory with durable contradictions and correction handles.
StateFuse differs by making contradictions durable first-class memory objects for downstream agent reasoning, while keeping resolution projection-scoped.

\subsection{Agent Memory Systems and Evaluation Gaps}
RAG and agent frameworks such as ReAct, Toolformer, Reflexion, Generative Agents, and MemGPT emphasize retrieval quality, tool use, and long-horizon behavior \citep{lewis2020retrieval,packer2023memgpt,park2023generative,schick2023toolformer,shinn2023reflexion,yao2022react}.
Those systems are often evaluated on whether the agent can remember enough, retrieve enough, or plan effectively.
They are less often evaluated on what happens when concurrent beliefs disagree or when a later correction must invalidate a stale earlier claim without erasing it.

Recent surveys and benchmark efforts call out weak evaluation standards for memory and the need for stronger task-level evidence \citep{Hu2025MemoryAgentBench,Mialon2024GAIA,Tan2025MemBench,Yoran2024AssistantBench,Zhang2025TOISMemorySurvey}.
Our matched-information benchmark design is intended as a step in that direction:
it separates the effects of information preservation, conflict structuring, abstention policy, and overwrite-style collapse rather than treating all memory surfaces as equivalent.

\subsection{Comparative Matrix}
Table~\ref{tab:related-work} positions StateFuse across merge semantics, contradiction handling, provenance, replayability, and mutability boundaries.
The distinctive property is not ``CRDTs exist'' but the combination of:
\begin{itemize}[leftmargin=*]
  \item correction by exact or semantic handle,
  \item explicit conflict surfacing as a public contract,
  \item bounded projection authority,
  \item projection-equivalent compaction,
  \item authenticated sync containment.
\end{itemize}

\begin{table}[t]
\centering
\caption{Related-work matrix across merge and conflict-management dimensions.}
\label{tab:related-work}
\begingroup
\small
\setlength{\tabcolsep}{2pt}
\renewcommand{\arraystretch}{1.08}
\begin{tabular}{>{\raggedright\arraybackslash}p{2.8cm} >{\raggedright\arraybackslash}p{2.1cm} >{\raggedright\arraybackslash}p{2.4cm} >{\raggedright\arraybackslash}p{1.6cm} >{\raggedright\arraybackslash}p{2.3cm} >{\raggedright\arraybackslash}p{2.8cm}}
\toprule
\shortstack{System\\Family} & \shortstack{Merge\\Semantics} & \shortstack{Conflict\\Treatment} & Prov. & \shortstack{Det.\\Replay} & \shortstack{Resolver\\Boundary} \\
\midrule
Classical KV/overwrite memory & Overwrite & Erased/hidden & Low & No & Mutable \\
OpSet/JSON CRDT systems \cite{Kleppmann2018OpSets,KleppmannBeresford2017JSONCRDT} & CRDT op-set/JSON merge & Structural conflicts preserved & High & Yes & Built-in data-type policy \\
Bayou/optimistic replication \cite{Terry1995Bayou,SaitoShapiro2005OptimisticReplication} & History + app policy & App-defined reconciliation & Medium & Partial & Mutable tentative state \\
Event-sourced / CQRS systems & Append-only log + projections & Usually app-level & Medium & Yes & Projection-defined \\
Provenance-aware memory systems \cite{w3cprov} & Varies & Often implicit & High & Partial & Usually mutable \\
RAG and agent memory stacks \cite{lewis2020retrieval,packer2023memgpt} & Retrieval or manager policy & Usually overwritten/implicit & Medium & Limited & Mutable \\
Agent-memory evaluations \cite{Zhang2025TOISMemorySurvey,Tan2025MemBench,Hu2025MemoryAgentBench} & Evaluation taxonomy & N/A & N/A & N/A & Highlights benchmark gaps \\
\textbf{StateFuse (this work)} & \textbf{Op-set memory substrate} & \textbf{Explicit ConflictSet + semantic correction} & \textbf{High} & \textbf{Yes} & \textbf{Immutable base / mutable projection} \\
\bottomrule
\end{tabular}
\endgroup
\end{table}

\section{Method}
\label{sec:method}

\subsection{Design Goals}
StateFuse is designed for agent workloads with branching execution, uncertain observations, validated corrections, and replica-level concurrency.
The design goals are:
\begin{enumerate}[leftmargin=*]
  \item deterministic convergence across benign replicas,
  \item immutable, replayable history,
  \item explicit contradiction preservation,
  \item correction semantics that work both locally and across replicas,
  \item projection-time resolution without base-memory mutation,
  \item bounded storage through projection-equivalent compaction,
  \item minimal authenticated sync guarantees for signed claims and retractions.
\end{enumerate}

\subsection{Operations, Claim Handles, and Predicate Contracts}
StateFuse stores immutable operations over four data objects:
\begin{itemize}[leftmargin=*]
  \item \textbf{Evidence}: reference-like records with \texttt{evidence\_id}, pointer, and metadata.
  \item \textbf{Claim}: epistemic atoms keyed by \texttt{(namespace, subject, predicate)} with a value, confidence, timestamp, evidence links, provenance, and two identities.
  \item \textbf{Retraction}: a targeted invalidation of a prior claim instance or semantic claim handle.
  \item \textbf{Decision}: append-only planning or execution metadata that never changes truth state.
\end{itemize}

The operation vocabulary is:
\begin{itemize}[leftmargin=*]
  \item \texttt{EvidenceAdded(evidence)},
  \item \texttt{ClaimAdded(claim)},
  \item \texttt{ClaimRetracted(target\_claim\_id?, target\_claim\_ref?,}\\
    \texttt{reason, supersedes\_claim\_id?, supersedes\_claim\_ref?)},
  \item \texttt{DecisionAdded(decision)}.
\end{itemize}

Each claim carries both:
\begin{itemize}[leftmargin=*]
  \item \textbf{\texttt{claim\_id}}: an exact per-assertion identifier for local correction and provenance, and
  \item \textbf{\texttt{claim\_ref}}: a deterministic semantic handle derived from the claim key and predicate-governed value contract.
\end{itemize}
This distinction is deliberate.
\texttt{claim\_id} is the precise local edit target; \texttt{claim\_ref} is the stable cross-replica correction handle that remains usable even when the original writer's opaque identifier is unavailable.

Predicate behavior is governed by a deterministic contract registry.
For each predicate, the registry may specify:
\begin{itemize}[leftmargin=*]
  \item whether the predicate is functional or multi-valued,
  \item a deterministic normalization function \texttt{normalize(value)},
  \item a deterministic equality function \texttt{equal(left,right)},
  \item whether normalization should also be applied when deriving \texttt{claim\_ref}.
\end{itemize}
The admissibility rule is simple:
these functions must be deterministic and replica-invariant.
The implementation therefore includes a contract checker that can reject unstable normalization or equality behavior on sample inputs before a registry is accepted into evaluation or deployment.

\subsection{Merge and Materialization}
Let $O$ denote a finite set of immutable operations keyed by \texttt{op\_id}.
Replica merge is still plain set union:
\[
\mathrm{merge}(O_1, O_2) = O_1 \cup O_2.
\]
This is a standard OpSet-style join \citep{Kleppmann2018OpSets}; the novelty in StateFuse lies in deterministic interpretation rather than in the merge algebra itself.

Materialization transforms the op-set into a queryable memory state by:
\begin{enumerate}[leftmargin=*]
  \item indexing evidence by \texttt{evidence\_id},
  \item indexing claims by \texttt{claim\_id},
  \item computing effective \texttt{claim\_ref} values under the predicate contract,
  \item collecting retractions by both exact identifier and semantic handle,
  \item removing inactive claims,
  \item grouping surviving claims by \texttt{ClaimKey},
  \item emitting explicit \texttt{ConflictSet} objects for functional predicates with multiple distinct active values.
\end{enumerate}

The correction semantics are:
\begin{itemize}[leftmargin=*]
  \item \textbf{Exact correction:} if a retraction targets \texttt{claim\_id}, only that concrete assertion becomes inactive.
  \item \textbf{Semantic correction:} if a retraction targets \texttt{claim\_ref}, every claim instance with that semantic handle becomes inactive, including later-arriving claims.
  \item \textbf{Unseen-target no-resurrection:} a semantic-handle retraction arriving before its target claims still suppresses those claims once they appear in the merged op-set.
  \item \textbf{Active-only conflicts:} retracted claims never contribute candidates to a conflict set.
\end{itemize}

\begin{figure}[t]
\centering
\resizebox{\textwidth}{!}{%
\begin{tikzpicture}[
  node distance=1.0cm and 1.6cm,
  entity/.style={
    draw,
    rounded corners,
    align=left,
    fill=gray!8,
    font=\small,
    text width=4.8cm,
    inner sep=4pt
  },
  entitysmall/.style={
    draw,
    rounded corners,
    align=left,
    fill=gray!8,
    font=\small,
    text width=3.8cm,
    inner sep=4pt
  },
  rel/.style={-{Latex[length=2.3mm]}, thick},
  lab/.style={font=\scriptsize, fill=white, inner sep=1pt}
]
\node[entity] (e) {\textbf{Evidence}\\evidence\_id, pointer, metadata};
\node[entity, right=of e] (c) {\textbf{Claim}\\claim\_id, key, value, confidence\\timestamp, evidence\_ids, provenance};
\node[entitysmall, below=of c] (r) {\textbf{ClaimRetracted}\\target\_claim\_id, reason\\supersedes\_claim\_id?};
\node[entitysmall, below=of e] (d) {\textbf{Decision}\\decision\_id, scope, payload};
\node[entity, below=of r] (f) {\textbf{ConflictSet}\\conflict\_id, key, candidates\\distinct\_values, reason};
\node[entitysmall, right=of f] (z) {\textbf{Resolution}\\chosen\_claim\_id or null\\reason, confidence?};

\draw[rel] (e.east) -- node[lab, above]{supports} (c.west);
\draw[rel] (c.south) -- node[lab, right]{target} (r.north);
\draw[rel] (r.south) -- node[lab, right]{deactivates} (f.north);
\draw[rel] (c.south east) to[out=-65, in=95] node[lab, right]{group by key} (f.north east);
\draw[rel] (f.east) -- node[lab, above]{resolved by} (z.west);
\draw[rel, dashed] (d.south) -- ++(0,-3.25) -| node[lab, below, yshift=-1pt, pos=0.66]{task scope} (z.south);
\end{tikzpicture}
}
\caption{Core StateFuse entities and relations used during materialization and projection.}
\label{fig:data-model}
\end{figure}
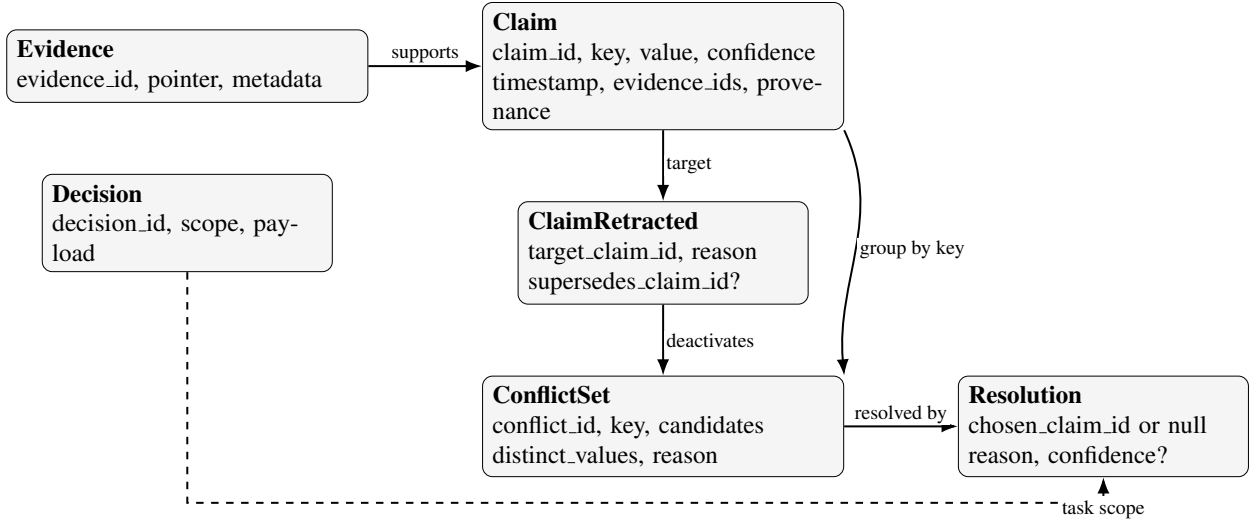

\subsection{Projection and Bounded Resolver Authority}
\texttt{build\_view(state, constraints, resolver)} derives a task-scoped projection from materialized memory.
It returns:
\begin{itemize}[leftmargin=*]
  \item selected claims when a key is conflict-free or the resolver chooses a candidate,
  \item unresolved conflict sets when the resolver abstains or fails validation,
  \item \texttt{surfaced\_conflicts}, the contradictions that actually reached the public decision surface,
  \item human-auditable explanations.
\end{itemize}

Resolvers operate only at projection time.
They may choose a candidate, abstain, or fail closed, but they cannot mutate base memory.
The conservative deterministic resolver used as the primary operational surface in this paper abstains on close or symmetric conflicts rather than forcing a confident collapse.
An optional LLM resolver is restricted to structured JSON I/O; malformed outputs, schema violations, and non-candidate selections remain unresolved and logged.

This makes \texttt{surfaced\_conflicts} a genuine contract boundary rather than a latent implementation detail.
The paper's contradiction metric is computed from what the surface actually exposes, not from hidden state that a baseline never shows to the decision policy.

\subsection{Bounded Operational Extensions}
Two implementation extensions are included for bounded validation but are not central to the main claim.
First, projection-equivalent compaction preserves the current resolver-visible semantics rather than the full historical log.
For a fixed predicate registry and deterministic resolver, the evaluation checks that compaction preserves active claims, conflict partitions, and projection outputs.
Second, authenticated sync adds a minimal signature-validation boundary that can reject revoked, malformed, or otherwise invalid incoming claims and retractions before merge.
These extensions are useful operationally, but in this paper they are treated as secondary support evidence rather than co-equal research contributions.

\subsection{Metrics}
Let $T$ be the set of tasks evaluated for a given surface.
We report:
\begin{itemize}[leftmargin=*]
  \item \textbf{Final accuracy}: fraction of all tasks in $T$ whose final prediction matches the task gold label, counting abstention as correct only on intentionally ambiguous tasks.
  \item \textbf{False certainty}: fraction of all tasks in $T$ on which the surface emits a non-null prediction that is incorrect.
  \item \textbf{Decision error rate}: fraction of non-abstaining predictions that are incorrect.
  \item \textbf{Surfaced conflict recall}: among tasks whose materialized state contains multiple semantically distinct active values, the fraction for which the surface explicitly exposes a contradiction.
  \item \textbf{Abstention quality}: among tasks whose gold label is abstention, the fraction for which the surface abstains.
  \item \textbf{Recoverability}: among correction tasks, the fraction for which the final surface recovers the corrected value.
\end{itemize}
These denominators are fixed in the benchmark code and reported explicitly to avoid the ambiguous or tautological interpretations that can arise when conflict presence is measured only from hidden internal state.

\subsection{Complexity}
With $|O|$ operations, materialization is linear in the size of the op-set plus the cost of sorting claims and retractions by key and identifier.
Conflict detection is linear in the number of active claims plus the work required by the predicate normalization/equality contract.
Projection is linear in the number of active keys and candidate sets actually surfaced to the resolver.
Projection-equivalent compaction is a full materialization pass followed by a filtered replay pass.

\subsection{Worked Example}
Consider two branches of the same agent task:
\begin{enumerate}[leftmargin=*]
  \item Branch A records \texttt{ClaimAdded(project, deadline,}\\
    \texttt{date=2026-03-25)}.
  \item Branch B independently records \texttt{ClaimAdded(project, deadline,}\\
    \texttt{date=2026-03-26)}.
  \item Materialization emits one \texttt{ConflictSet} over \texttt{project:deadline:date}.
  \item A downstream correction may retract either the exact stale assertion via \texttt{claim\_id} or the semantic stale handle via \texttt{claim\_ref}.
  \item A task-scoped projection may pick one candidate, abstain, or expose the conflict.
  \item If a planner commits a choice, it may append\\
    \texttt{DecisionAdded(scope=planning, payload=\{selected\_claim\_id: ...\})}\\
    without mutating the underlying claims.
\end{enumerate}

This is the intended StateFuse boundary:
history is immutable and auditable, correction is explicit, selection is projection-scoped, and bounded storage or authenticated sync do not alter the meaning of the surviving view.

\begin{figure}[t]
\centering
\resizebox{0.98\linewidth}{!}{%
\begin{tikzpicture}[
  node distance=0.9cm and 1.0cm,
  box/.style={
    draw,
    rounded corners,
    align=center,
    minimum height=1.0cm,
    fill=gray!8,
    font=\small,
    text width=3.1cm,
    inner sep=3pt
  },
  arr/.style={-{Latex[length=2.5mm]}, thick},
  lab/.style={font=\scriptsize, fill=white, inner sep=1pt}
]
\node[box] (api) {Memory API\\add evidence, add claim};
\node[box, right=of api] (store) {OpStore\\InMemory, JSONL, SQLite};
\node[box, right=of store] (oplog) {OpLog\\immutable operations};
\node[box, below=of oplog] (mat) {Materialize\\MemoryState};
\node[box, right=of mat] (conf) {ConflictSet\\detection};
\node[box, right=of conf] (resolver) {Resolver\\heuristic or LLM};
\node[box, right=of resolver] (view) {Projection\\selected claims + audit};

\draw[arr] (api.east) -- (store.west);
\draw[arr] (store.east) -- (oplog.west);
\draw[arr] (oplog.south) -- (mat.north);
\draw[arr] (mat.east) -- (conf.west);
\draw[arr] (conf.east) -- (resolver.west);
\draw[arr] (resolver.east) -- (view.west);
\draw[arr, dashed] (mat.north east) to[out=55, in=125] node[lab, above, yshift=2pt]{state + constraints} (resolver.north west);
\draw[arr, dashed] (view.north) to[out=130, in=10] node[lab, above]{base log unchanged} (oplog.east);
\end{tikzpicture}
}
\caption{StateFuse architecture: immutable merge substrate, explicit conflict materialization, and projection-time resolution.}
\label{fig:architecture}
\end{figure}
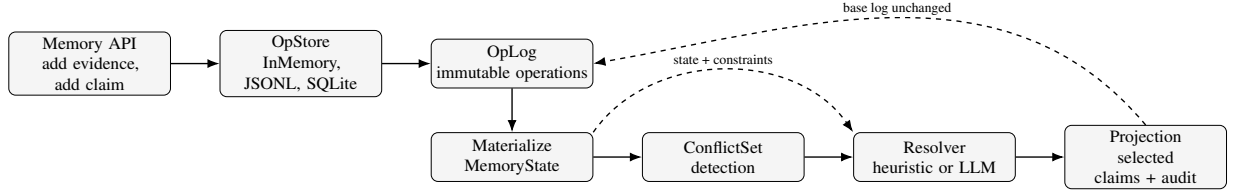

\begin{figure}[t]
\centering
\resizebox{\textwidth}{!}{%
\begin{tikzpicture}[
  life/.style={
    draw,
    rounded corners,
    minimum width=2.8cm,
    minimum height=0.8cm,
    align=center,
    fill=gray!8,
    font=\small
  },
  msg/.style={-{Latex[length=2.3mm]}, thick},
  note/.style={font=\scriptsize, align=center, fill=white, inner sep=1pt}
]
\node[life] (a) at (0,0) {Replica A};
\node[life] (b) at (3.3,0) {Replica B};
\node[life] (m) at (7.1,0) {Merger /\\Materializer};
\node[life] (v) at (10.9,0) {Projection};

\draw[densely dashed] (a.south) -- ++(0,-3.9);
\draw[densely dashed] (b.south) -- ++(0,-3.9);
\draw[densely dashed] (m.south) -- ++(0,-3.9);
\draw[densely dashed] (v.south) -- ++(0,-3.9);

\draw[msg] (0,-1.0) -- node[note, above]{ClaimAdded(op\_x)} (7.1,-1.0);
\draw[msg] (3.3,-1.8) -- node[note, above]{ClaimAdded(op\_y)} (7.1,-1.8);
\draw[msg] (7.1,-2.6) -- node[note, above]{merge = union(op\_x, op\_y)} (10.9,-2.6);
\draw[msg] (7.1,-3.4) -- node[note, above]{ConflictSet(key, candidates)} (10.9,-3.4);

\node[draw, rounded corners, fill=gray!8, align=left, font=\scriptsize, text width=3.6cm, anchor=north west]
  (n) at (9.0,-4.1) {Resolver chooses or abstains.\\Base op-log remains unchanged.};
\draw[msg, dashed] (10.9,-3.4) -- (n.north west);
\end{tikzpicture}
}
\caption{Branch-diverge-merge sequence with conflict-preserving materialization and projection-time resolution.}
\label{fig:sequence}
\end{figure}
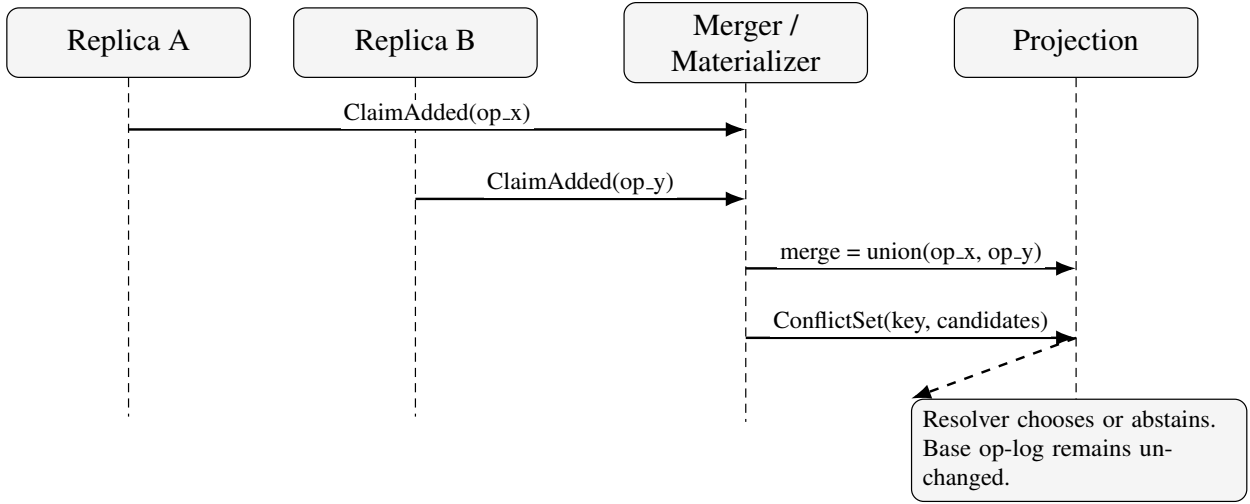

\section{Experiments}
\label{sec:experiments}

\subsection{Evaluation Structure}
The evaluation separates two evidence layers.
The main paper evidence is the official benchmark slice plus the uniform-verification agent loop.
The correction-handle ablation supports the semantics claim.

All manuscript aggregates are computed from run-level outputs.
When repeated runs are deterministic, the corresponding table captions say so once rather than repeating per-cell labels.

\begin{enumerate}[leftmargin=*]
  \item \textbf{Main empirical evidence}: official benchmark data and the uniform-verification agent loop.
  \item \textbf{Semantics support}: the correction-handle ablation.
\end{enumerate}

\subsection{Main Evidence: Official Benchmark and Uniform-Verification Agent Loop}
Table~\ref{tab:headline-results} summarizes the main paper-facing results.
Table~\ref{tab:memoryagentbench-official} gives the official benchmark comparison in full.
Table~\ref{tab:agent-loop} gives the downstream agent-loop comparison in full.

\begin{table}[t]
\centering
\caption{Headline practical results. Panel A reports the official MemoryAgentBench conflict-bearing slice under fair matched-information surfaces. Panel B reports the deterministic agent loop under a uniform verification budget. All repeated runs in Panel B are deterministic; confidence intervals are omitted.}
\label{tab:headline-results}
\begingroup
\small
\setlength{\tabcolsep}{4pt}
\renewcommand{\arraystretch}{1.08}
\begin{tabular}{>{\raggedright\arraybackslash}p{2.1cm} c c c c}
\toprule
\multicolumn{5}{c}{\textbf{Panel A: official benchmark}} \\
\midrule
Surface & Tasks & \shortstack[c]{Final\\Accuracy} & \shortstack[c]{Conflict\\Recall} & \shortstack[c]{False\\Certainty} \\
\midrule
StateFuse (cons.) & 282 & 64.9\% & 100.0\% & 2.1\% \\
StateFuse core & 282 & 64.9\% & 100.0\% & 2.1\% \\
MV register (cons.) & 282 & 64.9\% & 100.0\% & 2.1\% \\
Provenance surface (cons.) & 282 & 64.9\% & 100.0\% & 2.1\% \\
Raw log (cons.) & 282 & 64.9\% & 0.0\% & 2.1\% \\
Collapsed & 282 & 97.5\% & 0.0\% & 2.5\% \\
Oracle & 282 & 100.0\% & 100.0\% & 0.0\% \\
\bottomrule
\end{tabular}

\vspace{0.6em}

\begin{tabular}{>{\raggedright\arraybackslash}p{2.1cm} c c c c}
\toprule
\multicolumn{5}{c}{\textbf{Panel B: uniform-verification agent loop}} \\
\midrule
Surface & \shortstack[c]{Pre-Verify\\Success} & \shortstack[c]{Post-Verify\\Success} & \shortstack[c]{Post-Verify\\False Conf.} & \shortstack[c]{Avg Action\\Cost} \\
\midrule
StateFuse (cons.) & 80.0\% & 100.0\% & 0.0\% & 0.00 \\
StateFuse core & 80.0\% & 100.0\% & 0.0\% & 0.00 \\
MV register (cons.) & 80.0\% & 100.0\% & 0.0\% & 0.00 \\
Provenance surface (cons.) & 80.0\% & 100.0\% & 0.0\% & 0.00 \\
Raw log (cons.) & 80.0\% & 100.0\% & 0.0\% & 0.00 \\
Collapsed & 40.0\% & 60.0\% & 40.0\% & 2.20 \\
Oracle & 80.0\% & 100.0\% & 0.0\% & 0.00 \\
\bottomrule
\end{tabular}
\endgroup
\end{table}

\begin{table}[t]
\centering
\caption{Official MemoryAgentBench Conflict\_Resolution conflict-bearing subset under fair matched-information surfaces. Recoverability and abstention quality are N/A because this slice contains no correction-target or gold-abstention cases.}
\label{tab:memoryagentbench-official}
\begingroup
\small
\setlength{\tabcolsep}{4pt}
\renewcommand{\arraystretch}{1.08}
\begin{tabular}{>{\raggedright\arraybackslash}p{2.2cm} c c c c c c}
\toprule
Method & Tasks & \shortstack[c]{Final\\Accuracy} & \shortstack[c]{Conflict\\Recall} & \shortstack[c]{False\\Certainty} & \shortstack[c]{Recover-\\ability} & \shortstack[c]{Abstention\\Quality} \\
\midrule
StateFuse (cons.) & 282 & 64.9\% & 100.0\% & 2.1\% & N/A & N/A \\
StateFuse core & 282 & 64.9\% & 100.0\% & 2.1\% & N/A & N/A \\
MV register (cons.) & 282 & 64.9\% & 100.0\% & 2.1\% & N/A & N/A \\
Provenance surface (cons.) & 282 & 64.9\% & 100.0\% & 2.1\% & N/A & N/A \\
Raw log (cons.) & 282 & 64.9\% & 0.0\% & 2.1\% & N/A & N/A \\
StateFuse & 282 & 97.5\% & 100.0\% & 2.5\% & N/A & N/A \\
MV register & 282 & 97.5\% & 100.0\% & 2.5\% & N/A & N/A \\
Raw log & 282 & 97.5\% & 0.0\% & 2.5\% & N/A & N/A \\
Collapsed & 282 & 97.5\% & 0.0\% & 2.5\% & N/A & N/A \\
Oracle & 282 & 100.0\% & 100.0\% & 0.0\% & N/A & N/A \\
\bottomrule
\end{tabular}
\endgroup
\end{table}

\begin{table}[t]
\centering
\caption{End-to-end agent-loop benchmark with a uniform verification budget for every surface. All runs are deterministic; confidence intervals are omitted.}
\label{tab:agent-loop}
\begingroup
\scriptsize
\setlength{\tabcolsep}{2pt}
\renewcommand{\arraystretch}{1.08}
\begin{tabular}{>{\raggedright\arraybackslash}p{2.2cm} c c c c c c c c}
\toprule
Method & Tasks & \shortstack[c]{Pre-Verify\\Success} & \shortstack[c]{Post-Verify\\Success} & \shortstack[c]{Verification\\Gain} & \shortstack[c]{Pre-Verify\\False Conf.} & \shortstack[c]{Post-Verify\\False Conf.} & \shortstack[c]{Avg Tool\\Calls} & \shortstack[c]{Avg Action\\Cost} \\
\midrule
StateFuse (cons.) & 50 & 80.0\% & 100.0\% & 50.0\% & 20.0\% & 0.0\% & 2.40 & 0.00 \\
StateFuse core & 50 & 80.0\% & 100.0\% & 50.0\% & 20.0\% & 0.0\% & 2.40 & 0.00 \\
MV register (cons.) & 50 & 80.0\% & 100.0\% & 50.0\% & 20.0\% & 0.0\% & 2.40 & 0.00 \\
Provenance surface (cons.) & 50 & 80.0\% & 100.0\% & 50.0\% & 20.0\% & 0.0\% & 2.40 & 0.00 \\
Raw log (cons.) & 50 & 80.0\% & 100.0\% & 50.0\% & 20.0\% & 0.0\% & 2.40 & 0.00 \\
Collapsed & 50 & 40.0\% & 60.0\% & 50.0\% & 60.0\% & 40.0\% & 2.40 & 2.20 \\
Oracle & 50 & 80.0\% & 100.0\% & 50.0\% & 0.0\% & 0.0\% & 2.40 & 0.00 \\
StateFuse & 50 & 60.0\% & 80.0\% & 50.0\% & 40.0\% & 20.0\% & 2.40 & 1.20 \\
MV register & 50 & 60.0\% & 80.0\% & 50.0\% & 40.0\% & 20.0\% & 2.40 & 1.20 \\
Raw log & 50 & 60.0\% & 80.0\% & 50.0\% & 40.0\% & 20.0\% & 2.40 & 1.20 \\
\bottomrule
\end{tabular}
\endgroup
\end{table}

\textbf{Official benchmark result.}
The strongest external evidence is the converted official MemoryAgentBench \textit{Conflict\_Resolution} slice.
On the 282 conflict-bearing questions extracted from that release, StateFuse, flat multi-value, raw-log, and collapsed latest-write all reach 97.5\% final accuracy.
This means the official slice does \emph{not} support a claim that StateFuse is more accurate than strong flat or collapsed baselines in general.
What it does support is a surface-level distinction:
StateFuse and the flat conflict-preserving baselines expose contradictions on every conflict-bearing task in this slice, whereas raw-log and collapsed surfaces expose none.
Under conservative abstention, StateFuse, StateFuse core, flat multi-value, and provenance-style surfaces all move to the same operating point: 64.9\% accuracy, full contradiction recall, and 2.1\% false certainty.
That pattern shows that the main gain comes from public conflict surfacing plus abstention policy, not from a unique StateFuse answer-selection advantage.

\textbf{Uniform-verification agent-loop result.}
The agent loop is still synthetic, but it now removes the earlier surface-dependent tool-allocation confound.
Every method receives the same verification budget.
Under that policy, all conservative non-collapsing surfaces reach full post-verification success with no false-confident actions, while the collapsed surface reaches only 60\% success with 40\% false-confident actions.
This downstream result therefore supports a narrower claim:
preserving ambiguity and allowing conservative abstention is materially safer than collapsing memory before verification.
It does \emph{not} distinguish StateFuse from strong flat conservative baselines on this controlled task family.

\subsection{Correction-Handle Support: Exact vs.\ Semantic Correction Handles}
The main correction-handle evidence retained in the paper is the ablation in Table~\ref{tab:correction-handle}.

\begin{table}[t]
\centering
\caption{Correction-handle ablation across exact-target and semantic-target correction tasks. All runs are deterministic; confidence intervals are omitted.}
\label{tab:correction-handle}
\begingroup
\small
\setlength{\tabcolsep}{4pt}
\renewcommand{\arraystretch}{1.08}
\begin{tabular}{>{\raggedright\arraybackslash}p{2.4cm} c c c c c}
\toprule
\shortstack[c]{Handle\\Mode} & Tasks & \shortstack[c]{Final\\Accuracy} & \shortstack[c]{No\\Resurrection} & \shortstack[c]{ID-Target\\Projection Match} & \shortstack[c]{Semantic-Target\\Recoverability} \\
\midrule
Claim ID & 13 & 76.9\% & 76.9\% & 100.0\% & 0.0\% \\
Claim Ref & 13 & 100.0\% & 100.0\% & 100.0\% & 100.0\% \\
\bottomrule
\end{tabular}
\endgroup
\end{table}

\textbf{Result.}
The ablation includes semantic-only targets in addition to exact ID-targetable corrections.
On exact-target tasks, \texttt{claim\_ref} matches \texttt{claim\_id} projections exactly.
On semantic-target tasks, \texttt{claim\_ref} recovers the corrected value in every evaluated case, while \texttt{claim\_id} fails when the exact target identifier is unavailable.
In the same ablation, \texttt{claim\_ref} also avoids resurrection on all evaluated cases, whereas \texttt{claim\_id} drops to 76.9\%.
This is the clearest evidence that the semantic handle changes what corrections are expressible rather than merely renaming an exact retraction.

Taken together, the experiments support a narrow conclusion:
StateFuse is best supported as a safer public memory contract for contradiction surfacing, abstention, and semantic correction under fair matched-information comparisons.

\section{Discussion}
\label{sec:discussion}

\subsection{What Is Actually New Here}
The paper should not be read as claiming a new CRDT join.
Merge remains standard OpSet/CRDT machinery.
The novelty claim is instead at the contract layer:
\begin{itemize}[leftmargin=*]
  \item explicit public conflict objects,
  \item exact and semantic correction handles,
  \item deterministic predicate contracts,
  \item projection-bounded resolution,
  \item a matched-information evaluation protocol that keeps downstream policy budgets fixed.
\end{itemize}
That package is still partly compositional, but the correction-handle ablation shows that at least one piece---semantic correction via \texttt{claim\_ref}---changes which corrections are expressible.

\subsection{What the Strong Flat Baselines Mean}
The strongest empirical lesson is that a good flat baseline remains strong.
On the official MemoryAgentBench slice, StateFuse and the flat conflict-preserving baselines tie on both contradiction recall and, in the aggressive setting, answer accuracy.
Under conservative abstention, StateFuse, StateFuse core, flat multi-value, and provenance-style baselines again move together.
That matters because it sharpens the paper's claim.
The main question is not ``do explicit conflict objects always improve answer accuracy?''.
The main question is ``what public contract do we want a memory layer to expose when disagreement, correction, and abstention matter?''.

Our answer is that StateFuse is most compelling when:
\begin{itemize}[leftmargin=*]
  \item explicit surfaced contradiction is preferable to replaying a raw log,
  \item semantic correction targets matter across replicas or missing IDs,
  \item downstream policies must abstain safely rather than collapse silently,
  \item replay and audit requirements justify a stronger public contract than a flat candidate list.
\end{itemize}

\subsection{What the Current Results Do Not Justify}
The current evidence does \emph{not} justify a broad claim that StateFuse is generally more accurate than flat multi-value memory.
It also does not justify calling false-certainty or abstention metrics ``calibration'' in the formal probabilistic sense.
And it does not turn bounded compaction or authenticated sync mechanisms into definitive systems contributions on their own.
Those mechanisms are useful operational extensions, but they are not the core reason to accept the paper.

\subsection{Trust Boundary and Operational Scope}
StateFuse is still best understood as a bounded trust architecture.
Resolvers may influence projections but cannot rewrite replicated history.
The authenticated merge path can quarantine revoked, invalid, or malformed claims and retractions before they enter the op-set.
What it does \emph{not} provide is Byzantine fault tolerance, anti-spam economics, or authenticated replica membership beyond configured key policy.
Similarly, projection-equivalent compaction preserves the current resolver-visible view, not arbitrary future historical analysis.
These are useful operational guarantees, but bounded ones.

\subsection{Threats to Validity}
Several limits remain important:
\begin{itemize}[leftmargin=*]
  \item The main external evidence is one official benchmark slice rather than a broad benchmark portfolio.
  \item The paper does not yet report a large naturally arising trace study.
  \item The downstream agent loop and semantic-handle ablations are still controlled synthetic evaluations.
  \item Endpoint-specific live LLM comparisons are excluded from the main evidence.
  \item Compaction and authenticated sync are bounded validations, not full formal or adversarial evaluations.
\end{itemize}

These constraints should narrow the paper's position.
They do not remove the value of the contract design, but they do change the kind of claim the evidence can support.

\section{Conclusion}
\label{sec:conclusion}

StateFuse argues that agent-memory progress does not need to come from inventing a new replicated join.
It can also come from specifying a stronger public contract on top of standard immutable merge:
how contradictions are surfaced, how corrections are targeted, and what a downstream policy is allowed to see and decide.

The current results support a narrow conclusion.
StateFuse is not shown to be universally more accurate than strong flat multi-value baselines.
What the current results do support is that conflict-preserving surfaces are materially better than collapsed ones for contradiction surfacing, abstention, and safe post-verification behavior, and that semantic correction handles enable correction cases that exact IDs alone cannot express.

That is still a useful result.
It positions StateFuse less as a universal answer-selection win and more as a safer, more auditable memory contract for systems in which disagreement and correction are first-class operational concerns.
The next step is broader evidence on naturally arising public traces rather than additional in-repo stress tests.

\clearpage
\appendix
\section{Technical Appendix}
\label{sec:appendix-repro}

\subsection{Formal Objects}
Let $O$ be a finite set of immutable operations keyed by unique identifiers.
We partition operations by type:
\[
E(O),\ C(O),\ T(O),\ D(O)
\]
for evidence adds, claim adds, claim retractions, and decision adds respectively.

For a claim-add operation $c \in C(O)$, let:
\[
\mathrm{id}(c),\ \mathrm{ref}(c),\ \mathrm{key}(c),\ \mathrm{val}(c)
\]
denote its exact claim identifier, semantic claim handle, functional key, and value.
For a retraction $t \in T(O)$, let $\mathrm{target\_id}(t)$ and $\mathrm{target\_ref}(t)$ denote its optional exact and semantic targets.

We define the sets of exact and semantic tombstones as:
\[
R_{\mathrm{id}}(O) = \{ \mathrm{target\_id}(t) \mid t \in T(O),\ \mathrm{target\_id}(t)\neq\bot \},
\]
\[
R_{\mathrm{ref}}(O) = \{ \mathrm{target\_ref}(t) \mid t \in T(O),\ \mathrm{target\_ref}(t)\neq\bot \}.
\]
The set of active claims is:
\[
A(O) = \{ c \in C(O) \mid \mathrm{id}(c) \notin R_{\mathrm{id}}(O)\ \land\ \mathrm{ref}(c) \notin R_{\mathrm{ref}}(O) \}.
\]
This directly captures both exact targeted remove-wins semantics and semantic-handle unseen-target suppression:
if $\mathrm{id}(c) \in R_{\mathrm{id}}(O)$ or $\mathrm{ref}(c) \in R_{\mathrm{ref}}(O)$, then $c \notin A(O)$ regardless of arrival order.

Materialization is a deterministic interpretation function:
\[
M(O) = (\mathrm{Evidence}(O), \mathrm{Active}(O), \mathrm{Decisions}(O), \mathrm{Conflicts}(O)).
\]
Here $\mathrm{Active}(O)$ groups $A(O)$ by key, while $\mathrm{Conflicts}(O)$ contains one conflict object for every functional key whose active claims contain more than one distinct value under the predicate registry equality relation.

Projection is a separate function:
\[
V(M, \mathrm{resolver}, \mathrm{constraints}) = (\mathrm{Selected}, \mathrm{Unresolved}, \mathrm{Explanations}).
\]
The resolver may choose among candidates or abstain, but $V$ does not mutate $M$.

\subsection{Deterministic Interpretation Rules}
The implementation enforces deterministic traversal and tie-breaking:
\begin{itemize}[leftmargin=*]
  \item operations are traversed in sorted \texttt{op\_id} order,
  \item claim keys are traversed in sorted key order,
  \item candidates within a key are ordered by \texttt{claim\_id},
  \item conflict identifiers are derived from a canonical digest of the key and candidate identifiers,
  \item the default resolver uses a stable final tie-break on \texttt{claim\_id}.
\end{itemize}
These rules eliminate nondeterminism from hash-map iteration and ensure replayability for fixed inputs.

\subsection{Theorem Sketches}
\paragraph{Proposition 1 (Convergence).}
If two benign replicas hold the same op-set $O$, then their materialized states are equal:
\[
O_1 = O_2 \Rightarrow M(O_1) = M(O_2).
\]
\textit{Sketch.}
Materialization is a pure deterministic function of the op-set, and all intermediate traversals are stabilized by explicit sorting.

\paragraph{Proposition 2 (Deterministic Materialization).}
For any fixed finite op-set $O$, $M(O)$ is invariant to operation delivery order and merge order.
\textit{Sketch.}
The merge algebra is set union over immutable identifiers.
Set membership of operations is independent of delivery order, and all subsequent interpretation steps are deterministic functions of that set.

\paragraph{Proposition 3 (Retraction Semantics).}
If $c \in C(O)$ and there exists $t \in T(O)$ such that either $\mathrm{target\_id}(t)=\mathrm{id}(c)$ or $\mathrm{target\_ref}(t)=\mathrm{ref}(c)$, then $c \notin A(O)$.
This remains true even when $t$ is observed before $c$.
\textit{Sketch.}
The definition of $A(O)$ depends only on whether $\mathrm{id}(c)$ belongs to $R_{\mathrm{id}}(O)$ or $\mathrm{ref}(c)$ belongs to $R_{\mathrm{ref}}(O)$, not on causal or delivery order.

\paragraph{Proposition 4 (Conflict-Set Soundness).}
For every functional key $k$, if the active claims under $k$ contain more than one distinct value under the registry equality relation, then $\mathrm{Conflicts}(O)$ contains exactly one conflict object for $k$ whose candidates are exactly those active claims.
\textit{Sketch.}
Materialization first filters to active claims, then groups by key, and finally emits one conflict object per functional key when the distinct-value count exceeds one.
Retracted claims are excluded before conflict detection.

\paragraph{Proposition 5 (Projection Non-Interference).}
For any materialized state $M$ and resolver $r$, computing $V(M,r,\cdot)$ does not mutate $M$.
\textit{Sketch.}
Projection reads the materialized state and constructs a separate output object containing selected claims, unresolved conflicts, and explanations.
No update path writes back into the base memory structure.

\paragraph{Proposition 6 (Replayability).}
Given a fixed op-set $O$, deterministic predicate registry, deterministic resolver $r$, and fixed constraints $x$, the projection output is replayable:
\[
V(M(O), r, x)
\]
is identical across re-execution.
\textit{Sketch.}
This follows immediately from Propositions 1 and 2 plus deterministic resolver behavior.
For LLM-backed resolvers, replayability holds only when the resolver response itself is fixed and validated.

\subsection{Semantics Notes}
\begin{itemize}[leftmargin=*]
  \item \textbf{Retractions are targeted tombstones.} They may deactivate exact claim IDs or semantic claim handles, but not whole keys.
  \item \textbf{\texttt{supersedes\_claim\_id} is metadata only.} It may inform heuristics or provenance queries, but it does not modify truth semantics.
  \item \textbf{Decisions are planning metadata.} \texttt{DecisionAdded} contributes auditable scope-indexed records but does not enter conflict formation or claim activity.
  \item \textbf{Authenticated sync is a pre-merge boundary.} The propositions above apply to the accepted op-set after signature, revocation, timestamp, and collision checks quarantine invalid incoming traffic.
  \item \textbf{Invalid collisions are outside the benign model.} If the same \texttt{op\_id} appears with different payloads, strict merge raises an integrity error; checked sync quarantines the invalid operation and reports it explicitly.
\end{itemize}

\subsection{Complexity Notes}
Let $n = |O|$ be the number of operations and let $k$ be the number of distinct claim keys.
\begin{itemize}[leftmargin=*]
  \item \textbf{Merge:} near-linear in the number of inserted operations, plus deterministic iteration cost.
  \item \textbf{Materialization:} one pass for indexing plus sorting of keys and candidate lists; worst-case $O(n \log n)$.
  \item \textbf{Conflict detection:} for a key with $m$ active candidates, worst-case $O(m^2)$ distinct-value comparison under the predicate registry.
  \item \textbf{Projection:} linear in the number of keys plus candidate counts, excluding external resolver latency.
\end{itemize}
These bounds motivate future work on compaction, snapshotting, and larger-log benchmarks.

\clearpage
\bibliographystyle{plainnat}
\bibliography{references}

\end{document}